\documentclass[10pt,twocolumn,letterpaper]{article}
\usepackage{amsmath,amsfonts,bm}

\def\eqref#1{equation~\ref{#1}}
\def\1{\bm{1}}

\DeclareMathAlphabet{\mathsfit}{\encodingdefault}{\sfdefault}{m}{sl}
\SetMathAlphabet{\mathsfit}{bold}{\encodingdefault}{\sfdefault}{bx}{n}
\newcommand{\tens}[1]{\bm{\mathsfit{#1}}}

\def\tU{{\tens{U}}}
\def\tV{{\tens{V}}}
\def\tW{{\tens{W}}}

\newcommand{\R}{\mathbb{R}}

\newcommand{\Cov}{\mathrm{Cov}}
\usepackage{cvpr}
\usepackage{times}
\usepackage{epsfig}
\usepackage{graphicx}
\usepackage{amsmath}
\usepackage{amssymb}

\usepackage{url}
\usepackage{booktabs} 
\usepackage{subcaption}
\usepackage{eucal}
\usepackage{algorithm2e}
\usepackage{amsmath,bm}
\usepackage{mathabx}

\usepackage[pagebackref=true,breaklinks=true,letterpaper=true,colorlinks,bookmarks=false]{hyperref}

\cvprfinalcopy % *** Uncomment this line for the final submission

\def\cvprPaperID{****} % *** Enter the CVPR Paper ID here

\ifcvprfinal\fi
\begin{document}

%%%%%%%%% TITLE
\title{P-CapsNets, a General Form of Convolutional Neural Networks}

\author{
$\text{Zhenhua Chen}^{1}\text{, } \text{Xiwen Li}^{2}\text{, } \text{Chuhua Wang}^{1}\text{, } \text{David Crandall}^{1}$\\
$^{1}\text{Indiana University Bloomington}$\\
$^{2}\text{Washington University in St. Louis}$\\
{\tt\small \{chen478, cw234, djcran\}@indiana.edu}\\
{\tt\small \{xiwenli\}@wustl.edu}
}

% \and

% \and

\maketitle
%\thispagestyle{empty}

%%%%%%%%% ABSTRACT
\begin{abstract}
We propose Pure CapsNets (P-CapsNets) which is a generation of normal CNNs structurally. Specifically, we make three modifications to current CapsNets.  First, we remove routing procedures from CapsNets based on the observation that the coupling coefficients can be learned implicitly. Second, we replace the convolutional layers in CapsNets to improve efficiency. Third, we package the capsules into rank-3 tensors to further improve efficiency. The experiment shows that P-CapsNets achieve better performance than CapsNets with varied routing procedures by using significantly fewer parameters on MNIST\&CIFAR10. The high efficiency of P-CapsNets is even comparable to some deep compressing models. For example, we achieve more than 99\% percent accuracy on MNIST by using only 3888 parameters.  We visualize the capsules as well as the corresponding correlation matrix to show a possible way of initializing CapsNets in the future. We also explore the adversarial robustness of P-CapsNets compared to CNNs. 
\end{abstract}

%%%%%%%%% BODY TEXT
\section{Introduction}

Capsule Networks, or CapsNets, have been found to be more efficient for encoding the intrinsic spatial
relationships among features (parts or a whole) than normal CNNs. For
example, the CapsNet with dynamic routing~(\cite{dyrouting}) can
separate overlapping digits accurately, while the CapsNet with EM
routing~(\cite{emrouting}) achieves lower error rate on
smallNORB~(\cite{smallnorb}). However, the routing procedures of
CapsNets (including dynamic routing~(\cite{dyrouting}) and EM
routing~(\cite{emrouting})) are computationally expensive. Several
modified routing procedures have been proposed to improve the
efficiency~(\cite{fast_dyrouting, attention_routing, encapsule}), but
they sometimes do not ``behave as expected and often produce results
that are worse than simple baseline algorithms that assign the
connection strengths uniformly or randomly'' (\cite{bad_routing}). Another evidence comes from Hinton's recent work~\cite{capsule_unsupervised} which removes explicit routing procedures from capsule autoencoders.  

Even we can afford the computation cost of the routing procedures, we
still do not know whether the routing numbers we set for each layer
serve our optimization target. For example, in the work
of~\cite{dyrouting}, the CapsNet models achieve the best performance
when the routing number is set to 1 or 3, while other numbers cause
performance degradation. For a 10-layer CapsNet, assuming we have to
try three routing numbers for each layer, then $3^{10}$ combinations have
to be tested to find the best routing number assignment. This problem
could significantly limit the scalability and efficiency of CapsNets.

\begin {figure*}[t]
\begin{center} 
\includegraphics[width=0.6\textwidth]{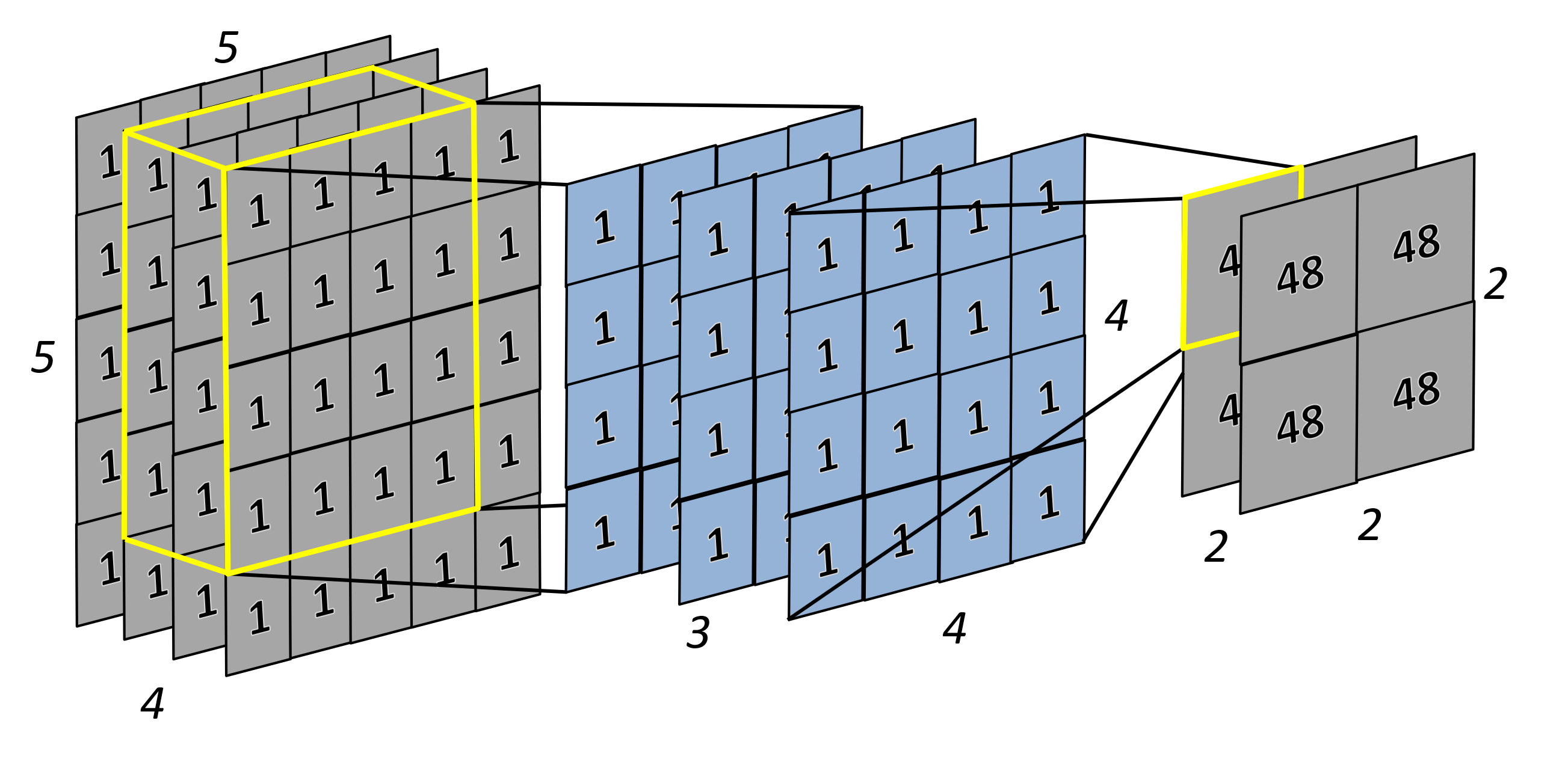}
\caption{3D convolution: tensor-to-scalar mapping. The shape of input is $5\times5\times4$. The shape of 3D kernel is $4\times4\times3$. As a result, the shape of output is $2\times2\times2$. Yellow area shows current input area being convolved by kernel and corresponding output.}
\label{3d_conv}
\includegraphics[width=.7\textwidth]{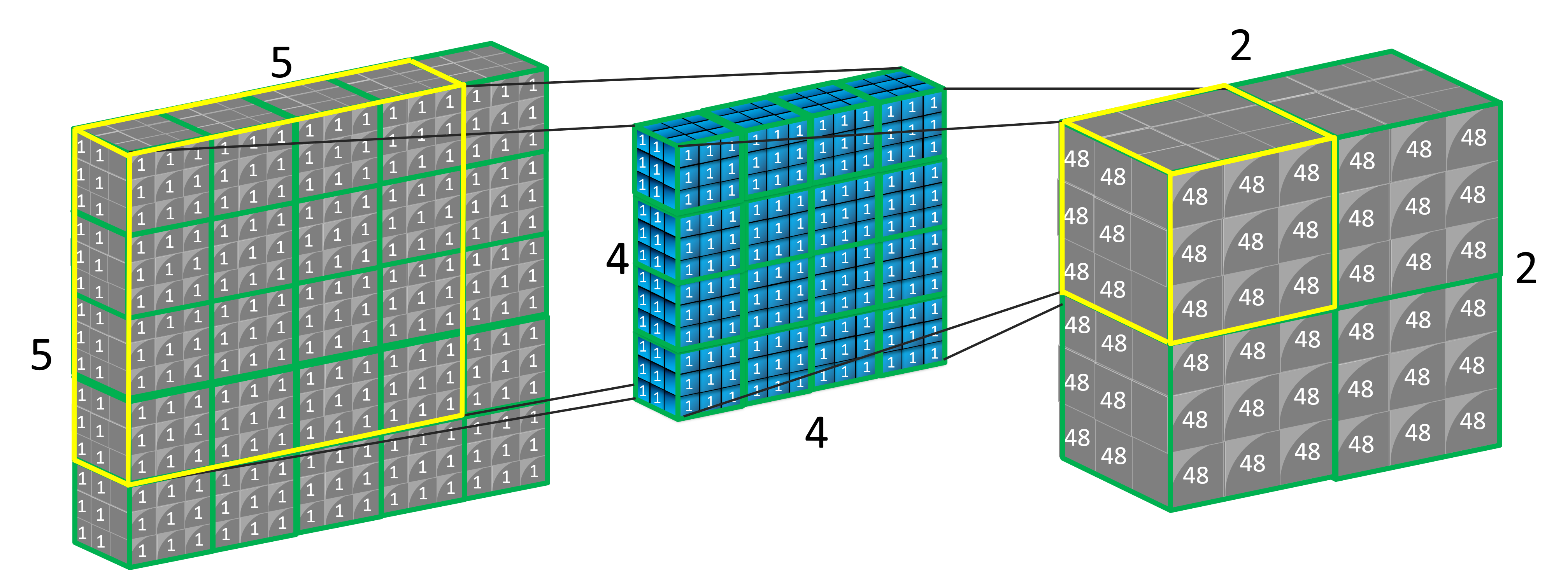}
\caption{Capsule convolution in P-CapsNets: tensor-to-tensor
  mapping. The input is a tensor of 1's which has a shape of $1\times
  5\times5\times(3\times3\times3)$ (correspond to the the input
  channel, input height, input width, first capsule dimension, 
  second capsule dimension, and third capsule dimension,
  respectively). The capsule kernel is also a tensor of 1's which has
  a shape of $4\times4\times1\times1\times(3\times3\times3)$ ---
  kernel height, kernel width, number of input channel, number of
  output channel, and the three dimensions of the 3D capsule. As a
  result, we get an output tensor of 48's which has a shape of
  $1\times2\times2\times(3\times3\times3)$. Yellow areas show current
  input area being convolved by kernel and corresponding output.}
\label{pcapsnet_structure} 
\end{center}
\end{figure*}

Here we propose
P-CapsNets, which resolve this issue by removing the routing procedures and instead
learning the coupling coefficients implicitly during
capsule transformation (see Section~\ref{main_idea} for
details). Moreover, another issue with current CapsNets is that it is common to
use several convolutional layers before feeding these
features into a capsule layer. We find that using convolutional layers
in CapsNets is not efficient, so we replace them with capsule
layers. Inspired by~\cite{emrouting}, we also explore how to package
the input of a CapsNet into rank-3 tensors to make P-CapsNets more
representative.
The capsule convolution in P-CapsNets can be considered as a more
general version of 3D convolution. At each step, 3D convolution uses a
3D kernel to map a 3D tensor into a scalar (as Figure~\ref{3d_conv}
shows) while the capsule convolution in
Figure~\ref{pcapsnet_structure} adopts a 5D kernel to map a 5D tensor
into a 5D tensor.

\section{Related Work}
% \djc{Probably need a short paragraph here describing the original CapsNet Hinton approach, i.e. explaining what a CapsNet is}
CapsNets~(\cite{dyrouting}) organize neurons as capsules to mimic the biological neural systems. One key design of CapsNets is the routing procedure which can combine lower-level features as higher-level features to better model hierarchical relationships. There have been many papers on improving the expensive routing
procedures since the idea of CapsNets was proposed. For example,
\cite{fast_dyrouting} improves the routing efficiency by 40\% by using
weighted kernel density estimation. \cite{attention_routing} propose
an attention-based routing procedure which can accelerate the dynamic
routing procedure. However, \cite{bad_routing} have found that these
routing procedures are heuristic and sometimes perform even worse
than random routing assignment.

Incorporating routing procedures into the optimization process could
be a solution. \cite{capsnet_optz} treats the routing procedure as a
regularizer to minimize the clustering loss between adjacent capsule
layers. \cite{encapsule} approximates the routing procedure with
master and aide interaction to ease the computation
burden. \cite{generalizedCaps} incorporates the routing procedure into
the training process to avoid the computational complexity of
dynamic routing.

Here we argue that from the viewpoint of optimization, the routing procedure,
which is designed to acquire coupling coefficients between adjacent
layers, can be learned and optimized implicitly, and may thus be unnecessary.
This approach is different from the above CapsNets which instead focus on improving the 
efficiency of the routing procedures, not attempting to replace them altogether.

\section{How P-CapsNets work}\label{main_idea}

We now describe our proposed P-CapsNet model in detail. We describe
the three key ideas in the next three sections: (1) that the routing
procedures may not be needed, (2) that packaging capsules into
higher-rank tensors is beneficial, and (3) that we do not need
convolutional layers.

\subsection{Routing procedures are not necessary}

The primary idea of routing procedures in CapsNets is to use the parts
and learned part-whole relationship to vote for 
objects. Intuitively, identifying an object by counting the votes
makes perfect sense. Mathematically, routing procedures can also be
considered as linear combinations of tensors. This is similar to the
convolution layers in CNNs in which the basic operation of a
convolutional layer is linear combinations (scaling and addition),
\begin{equation} \label{eq:0}
    v_j = \sum_{i} W_{ij} u_{i}.
\end{equation}
where $v_j$ is the $jth$ output scalar,
$u_i$ is the $ith$ input scalar, and $W_{ij}$ is the weight.

The case in CapsNets is a bit more complex since the dimensionalities of
input and output tensors between adjacent capsule layers are different and
we can not combine them directly. Thus we adopt a step to transform
input tensors ($\mathbf{u_{i}}$) into intermediate tensors
($\mathbf{u_{j|i}}$) by multiplying a matrix ($\mathbf{W_{ij}}$). Then
we assign each intermediate tensors ($\mathbf{u_{j|i}}$) a weight
$c_{ij}$, and now we can combine them together, 
\begin{equation} \label{eq:1}
    \mathbf{v_j} = \sum_{i}c_{ij} \mathbf{W_{ij}}\mathbf{u_{i}}.
\end{equation}%

where $c_{ij}$ are called coupling coefficients which are
usually acquired by a heuristic routing procedure~(\cite{dyrouting,
  emrouting}).

In conclusion, CNNs do linear combinations on scalars while CapsNets
do linear combinations on tensors. Using a routing procedure to
acquire linear coefficients makes  sense. However, if
Equation~\ref{eq:1} is rewritten as, 
\begin{equation} \label{eq:2}
    \mathbf{v_j} = \sum_{i}c_{ij} \mathbf{W_{ij}}\mathbf{u_{i}} = \sum_{i}\mathbf{W^{'}_{ij}}\mathbf{u_{i}}.
\end{equation}
then from the
viewpoint of optimization, it is not necessary to learn or calculate
$c_{ij}$ and $\mathbf{W_{ij}}$ separately since we can learn
$\mathbf{W^{'}_{ij}}$ instead. In other words, we can learn the
$c_{ij}$ implicitly by learning
$\mathbf{W^{'}_{ij}}$. Equation~\ref{eq:2} is the basic operation of
P-CapsNets only we extend it to the 3D case; please see
Section~\ref{packaging} for details.

By removing routing procedures, we no longer need an expensive step
for computing coupling coefficients. At the same time, we can
guarantee the learned $\mathbf{W^{'}_{ij}} = c_{ij} \mathbf{W_{ij}}$
is optimized to serve a target, while the good properties of CapsNets could
still be preserved (see section~\ref{exp} for details). We
conjecture that the strong modeling ability of CapsNets come from
this tensor to tensor mapping between adjacent capsule layers.

From the viewpoint of optimization, routing procedures do not
contribute a lot either. Taking the CapsNets in~(\cite{dyrouting}) as an
example, the number of parameters in the transformation operation is
$6\times 6\times 32 \times \left( 8 \times 16 \right) = 147,456$ while
the number of parameters in the routing operation equals to $6\times
6\times 32 \times 10=11,520$ --- the ``routing parameters" only represent
7.25\% of the total parameters and are thus negligible compared
to the ``transformation parameters." In other words, the benefit 
from routing procedures may be limited, even though they are the
computational bottleneck.

Equation~\ref{eq:0} and Equation~\ref{eq:2} have a similar form. We
argue that the ``dimension transformation" step of CapsNets can be
considered as a more general version of convolution. For example, if
each 3D tensor in P-CapsNets becomes a scalar, then P-CapsNets would
degrade to normal CNNs. As Figure~\ref{fig:cm_comp} shows, the basic
operation of 3D convolution is $\displaystyle f: \tU \in \R^{g\times
  m\times n} \rightarrow v \in \R$ while the basic operation of
P-CapsNet is $\displaystyle f: \tU \in \R^{g\times m\times n}
\rightarrow \tV \in \R^{g\times m\times p}$.

\subsection{Packaging capsules into higher rank tensors is helpful to save parameters}\label{packaging}

The capsules in~(\cite{dyrouting}) and~(\cite{emrouting}) are vectors
and matrices. For example, the capsules in~\cite{dyrouting}
have dimensionality $1152 \times 10 \times 8\times 16$ which can convert each
8-dimensional tensor in the lower layer into a 16-dimensional tensor
in the higher layer ($32 \times 6 \times 6 = 1152$ is the input number
and 10 is the output number). We need a total of $1152 \times 10 \times
8\times 16 = 1474560$ parameters. If we package each input/output
vector into $4\times 2$ and $4\times 4$ matrices, we need only $1152
\times 10 \times 2\times 4 = 92160$ parameters. This is the policy
adopted by~\cite{emrouting} in which 16-dimensional tensors are
converted into new 16-dimensional tensors by using $4\times 4$
tensors. In this way, the total number of parameters is reduced by a
factor of 15.

In this paper, the basic unit of input ($\tU \in \R^{g\times m\times
  n}$), output ($\tV \in \R^{g\times m\times p}$) and capsules ($\tW
\in \R^{g\times n\times p}$) are all rank-3 tensors. Assuming the
kernel size is ($\mathbf{kh} \times \mathbf{kw}$), the input capsule
number (equivalent to the number of input feature maps in CNNs) is
$\mathbf{in}$.  If we extend Equation~\ref{eq:2} to the 3D case, and
incorporate the convolution operation, then we obtain,
\begin{equation} \label{eq:3}
\begin{split}
   \tV =\left[\tV_{0, :, :}, \tV_{1, :, :}, \dots, \tV_{g, :, :}\right] = 
   \sum_{i}^{in}\sum_{j}^{kh}\sum_{k}^{kw} \\
   \left(\left[
    \tW_{0, i, j, k}\tU_{0, i, j, k}, \tW_{1, i, j, k}\tU_{1, i, j, k}, \dots, \tW_{g, i, j, k}\tU_{g, i, j, k} 
   \right] \right),
\end{split}
\end{equation}
which shows how to obtain an output tensor from input tensors in the
previous layer in P-CapsNets.

Assuming a P-CapsNet model is supposed to fit a function
$\displaystyle f: \R^{W\times H\times 3} \rightarrow \R$, the
ground-truth label is $y\in\R$ and the loss function
$\mathbf{\CMcal{L}} = ||f-y||$. Then in back-propagation, we calculate
the gradients with respect to the input $\tU$ and with respect to the 
capsules $\tW$,
\begin{equation} \label{eq:gradients1}
\begin{split}
\displaystyle \nabla_\tU \mathbf{\CMcal{L}} 
   =\left[\displaystyle \nabla_{\tU_{0, :, :}} \mathbf{\CMcal{L}}, \dots, \displaystyle \nabla_{\tU_{g, :, :}} \mathbf{\CMcal{L}} \right] =
   \sum_{i}^{in}\sum_{j}^{kh}\sum_{k}^{kw} \\
   \left( 
   \left[    \tW_{0, i, j, k}  \displaystyle \nabla_{\tV_{0, :, :}} \mathbf{\CMcal{L}}, \dots, \tW_{g, i, j, k}  \displaystyle \nabla_{\tV_{g, :, :}} \mathbf{\CMcal{L}} \right]
   \right),
\end{split}
\end{equation}
\begin{equation} \label{eq:gradients2}
\begin{split}
\displaystyle \nabla_\tW \mathbf{\CMcal{L}} 
   =\left[\displaystyle \nabla_{\tW_{0, :, :}} \mathbf{\CMcal{L}}, \dots, \displaystyle \nabla_{\tW_{g, :, :}} \mathbf{\CMcal{L}} \right] =
   \sum_{i}^{in}\sum_{j}^{kh}\sum_{k}^{kw} \\
   \left( 
   \left[    \tU_{0, i, j, k}  \displaystyle \nabla_{\tV_{0, :, :}} \mathbf{\CMcal{L}}, \dots, \tU_{g, i, j, k}  \displaystyle \nabla_{\tV_{g, :, :}} \mathbf{\CMcal{L}} \right]
   \right).
\end{split}
\end{equation}

The advantage of folding capsules into high-rank tensors is to reduce
the computational cost of dimension transformation between adjacent
capsule layers. For example, converting a $1\times 16$ tensor to
another $1\times 16$ tensor, we need $16\times 16 = 256$
parameters. In contrast, if we fold both input/output vectors to
three-dimensional tensors, for example, as $2\times 4\times 2$, then
we only need 16 parameters (the capsule shape is $2 \times 4 \times
2$). For the same number of parameters, folded capsules might be more
representative than unfolded ones. Figure~\ref{pcapsnet_structure}
shows what happens in one capsule layer of P-CapsNets in detail.

%Can we further dig the potential of CapsNet in this way? For example,
%if we package each input tensor into a rank-a tensor, then multiply it
%with a rank-b capsule, the output would be a rank-(a+b-2x) ($x\ge0$)
%tensor. We believe the efficiency of P-CapsNets can be further
%improved in this way. We will consider this idea in our future work.

\subsection{We can build a pure CapsNet without using any convolutional layers}

It is a common practice to embed convolutional layers in
CapsNets, which makes these CapsNets a hybrid network with both
convolutional and capsule layers~(\cite{dyrouting, emrouting,
  generalizedCaps}). One argument for using several convolutional
layers is to extract low level, multi-dimensional features.
We argue that this claim is not so persuasive
based on two observations, \textbf{1)}. The level of multi-dimensional
entities that a model needs cannot be known in advance, and it does
not matter, either, as long as the level serves our target;
\textbf{2)}. Even if a model needs a low level of multi-dimensional
entities, the capsule layer can still be used since it is a more
general version of a convolutional layer.

Based on the above observations, we build a ``pure" CapsNet by using
only capsule layers. One issue of P-CapsNets is how to process the
input if they are not high-rank tensors. Our solution is simply adding
new dimensions. For example, the first layer of a P-CapsNet can take
$1 \times w \times h \times (1\times 1\times 3) $ tensors as the input
(colored image), and take $1 \times w \times h \times (1\times 1\times
1)$ tensors as the input for gray-scale images. \\

In conclusion, P-CapsNets make three modifications over
CapsNets~(\cite{dyrouting}). First, we remove the routing procedures
from all the capsule layers. Second, we replace all the convolutional
layers with capsule layers. Third, we package all the capsules and
input/output as rank-3 tensors to save parameters. We keep the loss and
activation functions the same as in the previous work. Specifically, for each
capsule layer, we use the squash function $\tV = \left(1 -
\frac{1}{\mathbf{e}^{\|\tV\|}}\right) \frac{\tV}{\|\tV\|}$
in~(\cite{squash2}) as the activation function. We also use the same
margin loss function in~(\cite{dyrouting}) for classification tasks,
\begin{equation} \label{eq:4}
\begin{split}
\mathbf{\CMcal{L}_k} =\CMcal{T}_k * \max(0, \ m^+-\|\tV_k\|)^2 +  \\ \CMcal{\lambda} *  (1-\CMcal{T}_k) *  \max(0, \ \|\tV_k\| - m^-)^2,
\end{split}
\end{equation}

where $\CMcal{T}_k$ = 1 iff class k is present, and $m^{+} \ge 0.5$,
$m^{-} \ge 0$ are meta-parameters that represent the threshold for
positive and negative samples respectively. $\lambda$ is a weight that
adjust the loss contribution for negative samples.

\begin{figure}[!]
  \centering
  \includegraphics[width=1.0\linewidth]{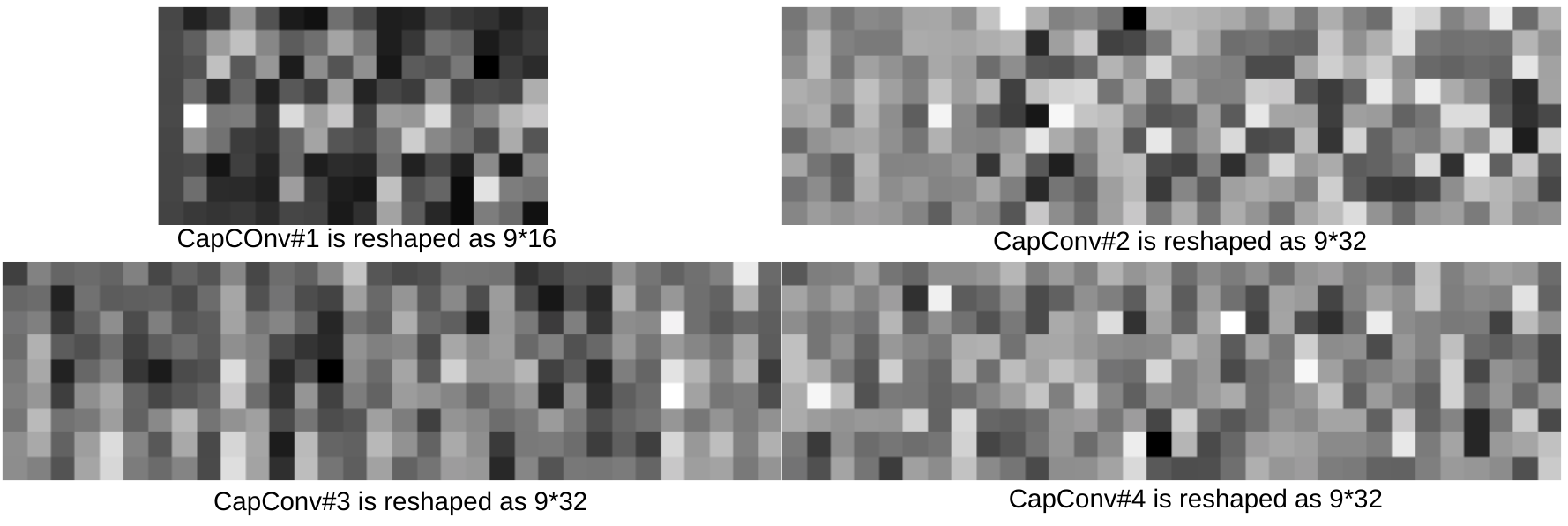}
  \caption{Visualization of filters in P-CapsNets. Top: conv1 layer, conv2 layer.  Bottom: conv3 layer, conv4 layer.}
  \label{pcapsnet_vis}
\end{figure}

\section{Experiments}\label{exp}
We test our P-CapsNets model on MNIST and CIFAR10. P-CapsNets show
higher efficiency than CapsNets~\cite{dyrouting} with various routing
procedures as well as several deep compressing neural network
models~\cite{compressedCNN, compress_1, compress_2}.

For MNIST, P-CapsNets\#0 achieve better performance than
CapsNets~\cite{dyrouting} by using 40 times fewer parameters, as
Table~\ref{tab:1} shows. At the same time, P-CapsNets\#3 achieve
better performance than Matrix CapsNets~\cite{emrouting} by using 87\%
fewer parameters. \cite{dense_dyrouting} is the only model that
outperforms P-CapsNets, but  uses 80 times more parameters.

Since P-CapsNets show high efficiency, it is interesting to
compare P-CapsNets with some deep compressing models on MNIST. We
choose five models that come from three algorithms as our
baselines. As Table~\ref{tab:2} shows, for the same number of
parameter, P-CapsNets can always achieve a lower error rate. For
example, P-CapsNets\#2 achieves 99.15\% accuracy by using only 3,888
parameters while the model~(\cite{compressedCNN}) achieves 98.44\% by
using 3,554 parameters. For P-CapsNet structures in Table~\ref{tab:1}
and Table~\ref{tab:2}, please check our supplementary materials 
for details.

\begin {table}[!]
\begin{center} 
\begin{tabular}{ lccc } 
\toprule
Models & routing & Error rate(\%) & Param \# \\
\midrule
DCNet++~(\cite{dense_dyrouting}) & Dynamic (-) & 0.29 & 13.4M \\ 
DCNet~(\cite{dense_dyrouting}) & Dynamic (-) & \boldmath{0.25} & 11.8M \\ 
CapsNets~(\cite{dyrouting}) & Dynamic (1) & $0.34_{\pm 0.03}$   & 6.8M \\
CapsNets~(\cite{dyrouting}) & Dynamic (3) & $0.35_{\pm 0.04}$   & 6.8M \\
Atten-Caps~(\cite{attention_routing} & Attention (-) & $0.64$ & $\approx$ 5.3M\\
CapsNets~(\cite{emrouting}) & EM (3) & $0.44$   & 320K \\
P-CapsNets\#0 & - & $0.32_{\pm0.03}$ & 171K\\
P-CapsNets\#3 & - & $0.41_{\pm0.05}$ & \boldmath{22.2K}\\
\bottomrule
\end{tabular}
\caption{Comparison between P-CapsNets and CapsNets with routing procedures in terms of error rate on MNIST. The number in each routing type means the number of routing times.}
\label{tab:1} 
\end{center}
\end{table}

\begin {table}[!]
\begin{center} 
\begin{tabular}{ lcc } 
\toprule
Algorithm & Error rate(\%) & Param \# \\
\midrule
KFC-Combined~(\cite{compress_2}) & 0.57 & 52.5K\\
Adaptive Fastfood 2048~(\cite{compress_1}) & $0.73$ & 52.1K\\
% \textcolor{red}{P-CapsNets\#4} & \textcolor{red}{$0.48_{\pm0.04}$} & \textcolor{red}{44.1K}\\ 
Adaptive Fastfood 1024~(\cite{compress_1}) & $0.73$ & 38.8K\\
KFC-II~(\cite{compress_2}) & 0.76 & 27.7K\\
P-CapsNets\#3 & $0.41_{\pm0.05}$ & 22.2K\\ 
P-CapsNets\#2 & $0.85_{\pm0.08}$ & 3.8K\\ 
ProfSumNet~(\cite{compressedCNN}) & 1.55 & 3.6K\\
P-CapsNets\#1 & $1.05_{\pm0.07}$ & 2.9K\\ 
\bottomrule
\end{tabular}
\caption{
Comparison between P-CapsNets and three compressing algorithms in terms of error rate on MNIST.}
\label{tab:2} 
\end{center}
\end{table}

For CIFAR10, we also adopt a five-layer P-CapsNet (please see the
supplementary materials) which has about 365,000
parameters. We follow the work of~\cite{dyrouting, emrouting} to crop
24 $\times$ 24 patches from each image during training, and use
only the center 24 $\times$ 24 patch during testing. We also use
the same data augmentation trick as in~\cite{maxout} (please see
our supplementary materials for details). As Table~\ref{tab:3} shows, 
P-CapsNet achieves better performance than several routing-based
CapsNets by using fewer parameters. The only exception is
Capsule-VAE~(\cite{VB-Routing}) which uses fewer parameters than
P-CapsNets but the accuracy is lower. The structure of P-CapsNets\#4 can
be found in our supplementary materials.

In spite of the parameter-wise efficiency of P-CapsNets, one
limitation is that we cannot find an appropriate acceleration
solution like cuDNN~(\cite{cuDNN}) since all current acceleration
packages are convolution-based. To accelerate our training, we developed
a customized acceleration solution based on cuda~(\cite{cuda}) and
CAFFE~(\cite{caffe}). The primary idea is reducing the communication
times between CPUs and GPUs, and maximizing the number of
can-be-paralleled operations. Please check our supplementary materials
for details, and the code will be released soon.

\begin {table*}[!]
\begin{center} 
\begin{tabular}{ lcccc } 
\toprule
Models & Routing & Ensembled & Error rate(\%) & Param \# \\
\midrule
DCNet++~(\cite{dense_dyrouting}) & Dynamic (-) & 1 &10.29 & 13.4M \\ 
DCNet~(\cite{dense_dyrouting}) & Dynamic (-) & 1 &18.37 & 11.8M \\ 
MS-Caps~(\cite{ms_caps}) & Dynamic (-) & 1 &24.3 & 11.2M \\ 
CapsNets~(\cite{dyrouting}) & Dynamic (3) & 7 & $10.6$   & 6.8M \\
Atten-Caps~(\cite{attention_routing} & Attention (-)& 1 & $11.39$ & $\approx$5.6M\\
FRMS~(\cite{fast_dyrouting}) & Fast Dynamic (2) & 1 & $15.6$&  1.2M \\  
FREM~(\cite{fast_dyrouting}) & Fast Dynamic (2) & 1 & $14.3$&  1.2M \\  
CapsNets~(\cite{emrouting}) & EM (3) & 1 & $11.9$   & 458K \\
P-CapsNets\#4 & - & 1 & $10.03$ & 365K\\
Capsule-VAE~(\cite{VB-Routing}) & VB-Routing & 1 & 11.2 & \boldmath{323K}\\

% \textcolor{red}{P-CapsNets} & - & 1 & \textcolor{red}{$0$} & \textcolor{red}{0K}\\
\bottomrule
\end{tabular}
\caption{Comparison between P-CapsNets and several CapsNets with routing procedures in terms of error rate on CIFAR10. The number in each routing type is the number of routing times.}
\label{tab:3} 
\end{center}
\end{table*}

\section{Visualization of P-CapsNets}
We visualize the capsules (filters) of P-CapsNets trained on MNIST
(the model used is the same as in
Figure~\ref{appendix:pcapsnet_eg}). The capsules in each layer are 
7D tensors. We flatten each layer into a matrix to make it easier to
visualize. For example, the first capsule layer has a shape of
$3\times3\times 1\times 1 \times (1 \times1\times 16)$, so we reshape it
to a $9\times 16$ matrix. We do a similar reshaping for the following
three layers, and the result is shown in Figure~\ref{pcapsnet_vis}.

\begin{figure*}[!]
\centering
\begin{tabular}{@{}cccc@{}}
\multicolumn{4}{c}{\includegraphics[width=0.3\textwidth]{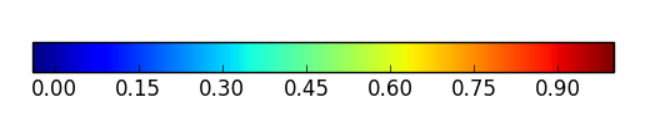}}\\
\includegraphics[width=.23\textwidth]{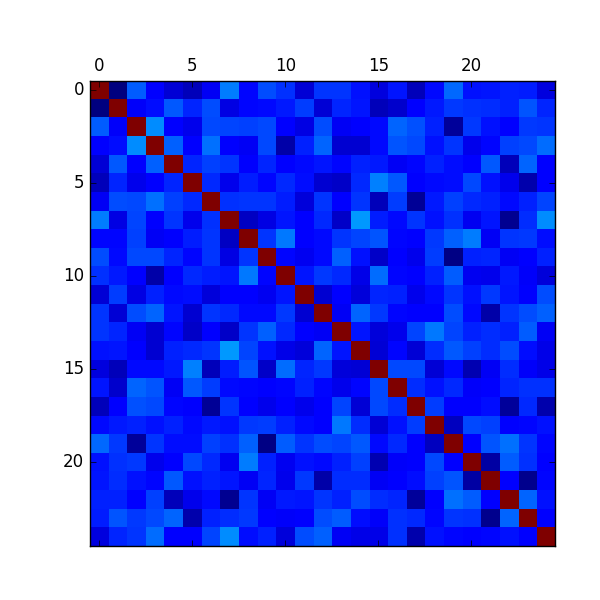} &
\includegraphics[width=.23\textwidth]{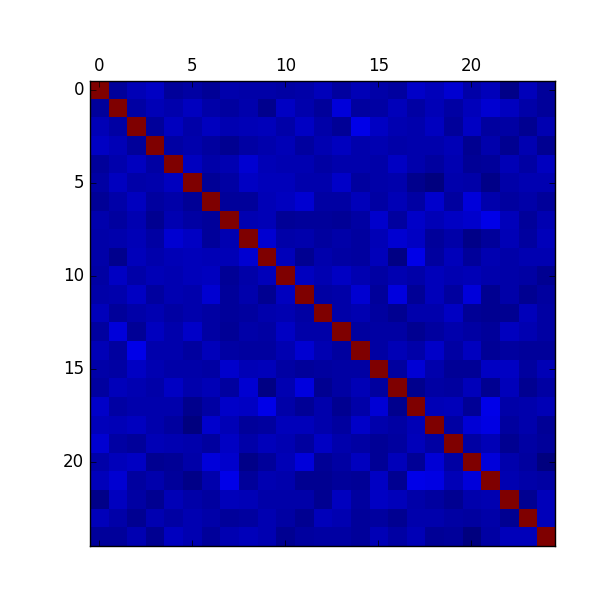} &
\includegraphics[width=.23\textwidth]{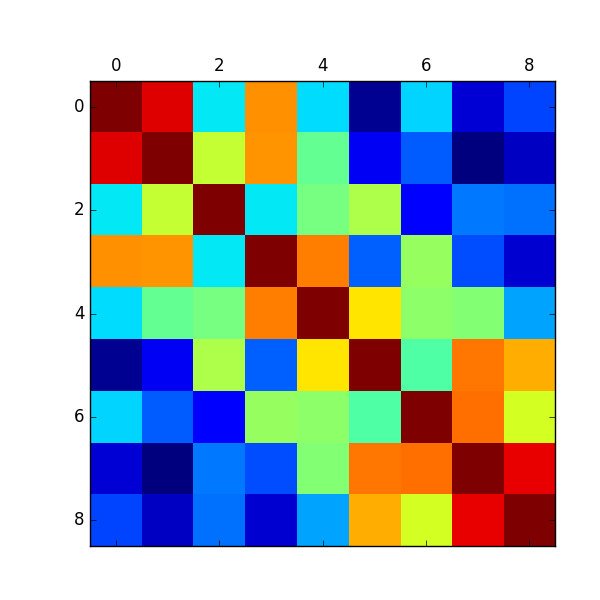} &
\includegraphics[width=.23\textwidth]{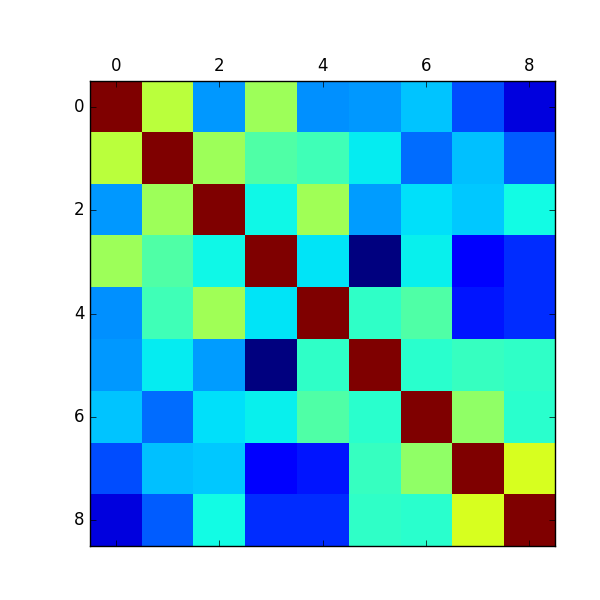}\\
(a) conv1&
(b) conv2&

(c) capconv1&
(d) capconv2
 
\end{tabular}

\caption{Correlation Matrix of Convolution Layers between CNNs and P-CapsNets. CNN: (a) conv1: 5x5x1x256; (b) conv2: 5x5x256x256. P-CaspNets: (c) capconv1: 1x1x3x3x(1x1x16); (d) capconv2: 1x1x3x3x(1x4x8).}
\label{fig:cm_comp}
\end{figure*}

We observe that the capsules within each layer appear correlated with
each other. To check if this is true, we print out the first two
layers' correlation matrix for both the P-CapsNet model as well as a
CNN model (which comes from \cite{dyrouting}, also trained on MNIST)
for comparison. We compute Pearson product-moment correlation
coefficients (a division of covariance matrix and multiplication of
standard deviation) of filter elements in each of two convolution
layers respectively. In our case, we draw two 25x25 correlation
matrices from that reshaped conv1 (25x256) and conv2
(25x65536). Similarly, we generate two 9x9 correlation matrices of
P-CapsNets from reshaped conv1 (9x16) and conv2 (9x32). As Figure
\ref{fig:cm_comp} shows, the filters of convolutional layers have
lower correlations within kernels than P-CapsNet. The result makes
sense since the capsules in P-CapsNets are supposed to extract the
same type of features while the filters in standard CNNs are supposed
to extract different ones.

The difference shown here suggests that we might rethink  the initialization of
CapsNets. Currently, our P-CapsNet, as well as other types of CaspNets
all adopt initializing methods designed for CNNs, which might not be ideal.

\begin{figure}[!]
  \centering
  \includegraphics[width=0.7\linewidth]{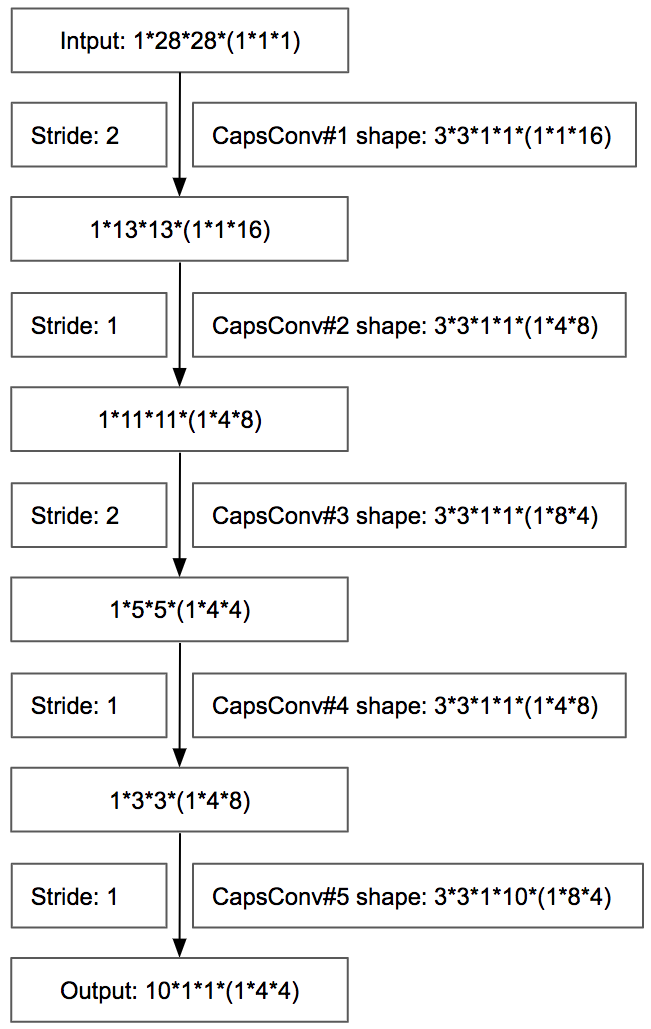}
  \caption{The structure P-CapsNets\#2. The input are gray-scale images with a shape of 28 $\times$ 28, we reshape it as a 6D tensor, $1\times28\times28\times (1\times1\times 1)$ to fit our P-CaspNets. The first capsule layer (CapsConv\#1, as Figure~\ref{appendix:pcapsnet_eg} shows.), is a 7D tensor,  $3\times3\times1\times 1\times (1\times1\times 16)$. Each dimension of the 7D tensor represents the kernel height, the kernel width, the number of input capsule feature map, the number of output capsule feature map, the capsule's first dimension, the capsule's second dimension, the capsule's third dimension. All the following feature maps and filters can be interpreted in a similar way.}
  \label{appendix:pcapsnet_eg}
\end{figure}

\begin{figure}[!]
\centering
\begin{minipage}{.5\textwidth}
  \centering
  \includegraphics[width=1\linewidth]{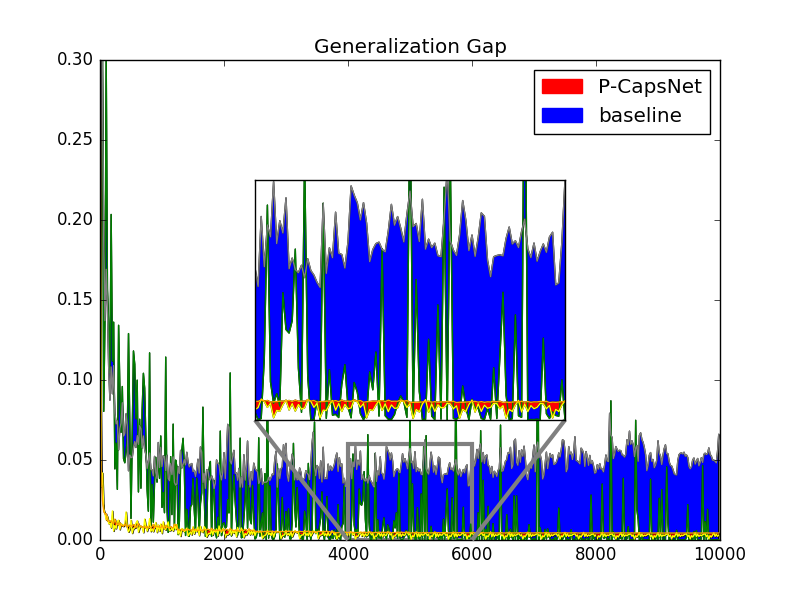}
  \captionof{figure}{Generalization Gap of P-CapsNets (blue area) and the baseline (red area).}
  \label{figure:gap}
\end{minipage}%

\end{figure}

\section{Generalization Gap}
Generalization gap is the difference between a model's performance on
training data and that on unseen data from the same distribution. We
compare the generalization gap of P-CapsNets with that of the CNN
baseline~\cite{dyrouting} by marking out an area between training loss
curve and testing loss curve, as Figure \ref{figure:gap} shows. For
visual comparison, we draw the curve per 20 iterations for baseline
\cite{dyrouting} and 80 iterations for P-CapsNet, respectively. We can
see that at the end of the training, the gap of training/testing loss
of P-CapsNets is smaller than the CNN model. We conjecture that
P-CapsNets have a better generalization ability.

\begin{figure}[!]
\centering
\begin{minipage}{.5\textwidth}
  \centering
  \includegraphics[width=1\linewidth]{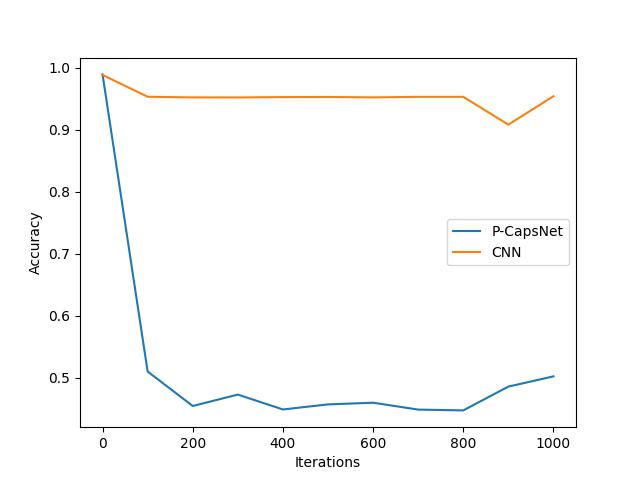}
  \captionof{figure}{P-CapsNets versus CNNs on white-box adversarial attack.}
  \label{figure:whitebox}
\end{minipage}
\end{figure}

\section{Adversarial Robustness}
For black-box adversarial attack, \cite{emrouting} claims that
CapsNets is as vulnerable as CNNs. We find that P-CapsNets also suffer
this issue, even more seriously than CNN models. Specifically, we
adopt FGSM~\cite{fgsm} as the attacking method and use LeNet as the
substitute model to generate one thousand testing adversarial
images. As~Table~\ref{tab:white-box} shows, when epsilon increases
from 0.05 to 0.3, the accuracy of the baseline and the P-CapsNet model
fall to 54.51\% and 25.11\%, respectively.

\begin {table}[!]
\begin{center} 
\begin{tabular}{ lcc } 
\toprule
Epsilon & \textbf{Baseline} & \textbf{P-CapsNets}  \\ 
\midrule
0.05 & 99.09\% & 98.66\% \\
0.1 &  98.01\%  & 94.4\% \\
0.15 & 95.52\% & 81.35\%\\ 
0.2 & 89.84\% & 59.52\%\\
0.25 & 78.31\% & 39.58\%\\
0.3 & 54.51\% & 25.11\%\\
\bottomrule
\end{tabular}
\caption{Robustness of P-CapsNets. The attacking method is FGSM~\cite{fgsm}. In this table, we use The baseline is the same CNN model in~\cite{dyrouting}.}
\label{tab:white-box} 
\end{center}
\end{table}

\cite{emrouting} claims that CapsNets show far more resistance to
white-box attack; we find an opposite result for
P-CapsNets. Specifically, we use
UAP~(\cite{universal_adversarial_perturbations}) as our attacking
method, and train a generative network (see
the supplementary materials for details) to generate universal
perturbations to attack the CNN model~(\cite{dyrouting}) as well as
the P-CapsNet model shown in Figure~\ref{appendix:pcapsnet_eg}). The
universal perturbations are supposed to fool a model that predicts a
targeted wrong label ((the ground truth label + 1) \% 10). As
Figure~\ref{figure:whitebox} shows, when attacked, the accuracy of the
P-CapsNet model decreases more sharply than the baseline.

It thus appears that P-CapsNets are more vulnerable to both white-box and
black-box adversarial attacking compared to CNNs. One possible reason
is that the P-CapsNets model we use here is significantly smaller than
the CNN baseline (3688 versus 35.4M). It would be a fairer comparison
if two models have a similar number of parameters.

\section{Conclusion}
We propose P-CapsNets by making three modifications based on
CapsNets~\cite{dyrouting}, 1) We replace all the convolutional layers
with capsule layers, 2) We remove routing procedures from the whole
network, and 3) We package capsules into rank-3 tensors to further
improve the efficiency. In this way, P-CapsNets becomes a general version of CNNs structurally. The experiment shows that P-CapsNets can achieve better performance than multiple other CapsNets variants with different routing procedures, as well as than deep compressing models, by using fewer parameters. We visualize the capsules in P-CapsNets and point out that the initializing methods of CNNs might not be appropriate for CapsNets. We conclude that the capsule layers in P-CapsNets can be considered as a general version of 3D convolutional layers.  We conjecture that CapsNets can encode the intrinsic spatial relationship between a part and a while efficiently, comes from the
tensor-to-tensor mapping between adjacent capsule layers. This mapping is presumably also the reason for P-CapsNets' good performance.

\section{Future work}
Apart from high efficiency, another advantage of CapsNets is extracting good spatial features. P-CapsNets have shown high efficiency in classification tasks, and should also be able to generalize well on segmentation \& detection tasks. This will be our feature work.

{\small
\bibliographystyle{ieee_fullname}
\bibliography{main}
}

\clearpage

\appendix
\section{Network Structures}\label{appendix:structures}
\subsection{MNIST\&CIFAR}
For MNIST\&CIFAR10, we designed five versions of CapsNets (CapsNets\#0, CapsNets\#1, CapsNets\#2, CapsNets\#3), they are all five-layer CapsNets. Take CapsNets\#2 as an example, the input are gray-scale images with a shape of 28 $\times$ 28, we reshape it as a 6D tensor, $1\times28\times28\times (1\times1\times 1)$ to fit our P-CaspNets. The first capsule layer (CapsConv\#1, as Figure~\ref{appendix:pcapsnet_eg} shows.), is a 7D tensor,  $3\times3\times1\times 1\times (1\times1\times 16)$. Each dimension of the 7D tensor represents the kernel height, the kernel width, the number of input capsule feature map, the number of output capsule feature map, the capsule's first dimension, the capsule's second dimension, the capsule's third dimension. All the following feature maps and filters can be interpreted in a similar way. 
\begin{figure}[!]
  \centering
  \includegraphics[width=0.5\linewidth]{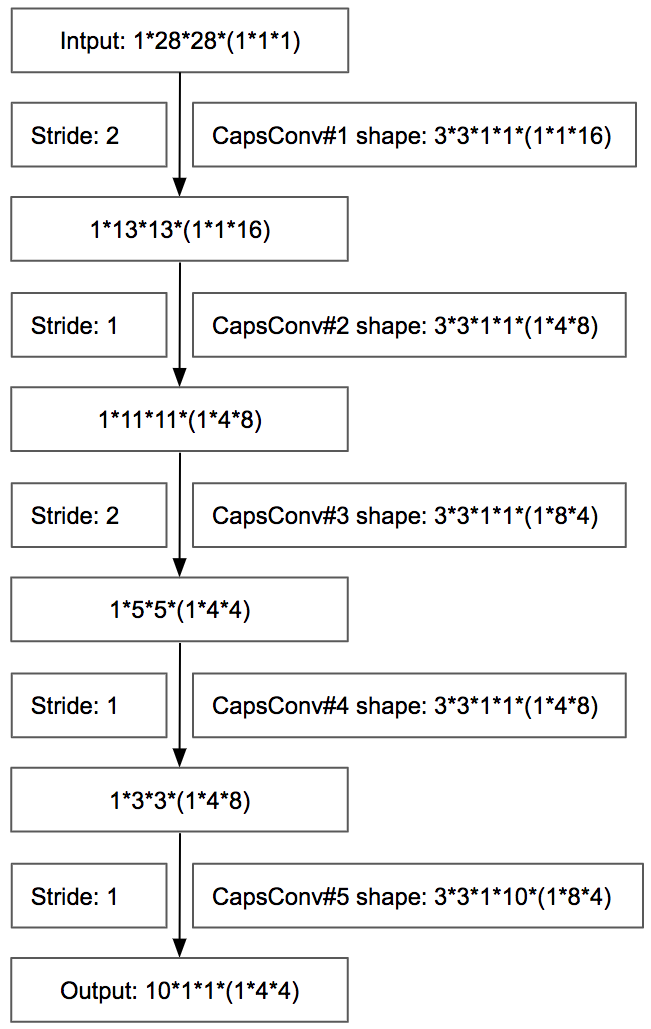}
  \caption{The structure P-CapsNets\#2. }
  \label{appendix:pcapsnet_eg}
\end{figure}

Similarly, the five capsule layers of P-CapsNets\#0 are $3\times3\times 1\times 1\times (1\times 1\times 32$, $3\times3\times 1\times 2 \times (1\times8\times 8)$, $3\times3\times 2\times 4 \times (1 \times 8\times 8)$, $3\times3\times 4\times 2 \times (1\times 8\times 8$, $3\times 3 \times 2\times 10 \times (1 \times 8\times 8)$ respectively. The strides for each layers are $(2, 1, 2, 1, 1)$.

The five capsule layers of P-CapsNets\#1 are $3\times3\times 1\times 1\times (1\times 1\times 16$, $3\times3\times 1\times 1 \times (1\times4\times 6)$, $3\times3\times 1\times 1 \times (1 \times 6\times 4)$, $3\times3\times 1\times 1 \times (1\times 4\times 6$, $3\times 3 \times 1\times 10 \times (1 \times 6\times 4)$ respectively. The strides for each layers are $(2, 1, 2, 1, 1)$.

The five capsule layers of P-CapsNets\#3 are $3\times3\times 1\times 1\times (1\times 1\times 32$, $3\times3\times 1\times 4 \times (1\times8\times 16)$, $3\times3\times 4\times 8 \times (1 \times 16\times 8)$, $3\times3\times 8\times 4 \times (1\times 8\times 16$, $3\times 3 \times 4\times 10 \times (1 \times16\times 16)$ respectively. The strides for each layers are $(2, 1, 2, 1, 1)$.

The five capsule layers of P-CapsNets\#4 are $3\times3\times 1\times 1\times (1\times 3\times 32$, $3\times3\times 1\times 4 \times (1\times8\times 16)$, $3\times3\times 4\times 8 \times (1 \times 16\times 8)$, $3\times3\times 8\times 10 \times (1\times 8\times 16$, $3\times 3 \times 10\times 10 \times (1 \times16\times 16)$ respectively. The strides for each layers are $(2, 1, 1, 2, 1)$.

\subsection{The Generative Network for Adversarial Attack}\label{gen_network}
The input of the generative network is a 100-dimension vector filled with a random number ranging from -1 to 1. Then the vector is fed to a fully-connected layer with 3456 output ( the output is reshaped as $3\times 3\times 384$). On top of the fully-connected layer, there are three deconvolutional layers. They are  one deconvolutional layer  with 192 output (the kernel size is 5, the stride is 1, no padding), one deconvolutional layer with 96 output  (the kernel size is 4, the stride is 2, the padding size is 1), and one deconvolutional layer with 1 output  (the kernel size is 4, the stride is 2, the padding size is 1) respectively. The final output of the three deconvolutional layers has the same shape as the input image (28$\times$28) which are the perturbations. 

\section{Meta-parameters \& Data Augmentation} \label{data_aug}
For all the P-CapsNet models in the paper, We add a Leaky ReLU function(the negative slope is 0.1) and a squash function after each capsule layer. All the parameters are initialized by MSRA~(\cite{msra}).

For MNIST, we decrease the learning rate from 0.002 every 4000 steps by a factor of 0.5 during training. The batch size is 128, and we trained our model for 30 thousand iterations. The upper/lower bound of the margin loss is 0.5/0.1. The $\lambda$ is 0.5. We adopt the same data augmentation as in ~(\cite{dyrouting}), namely, shifting each image by up to 2 pixels in each direction with zero padding.

For CIFAR10, we use a batch size of 256. The learning rate is 0.001, and we decrease it by a factor of 0.5 every 10 thousand iterations. We train our model for 50 thousand iterations. The upper/lower bound of the margin loss is 0.6/0.1. The $\lambda$ is 0.5. Before training we first process each image by using Global Contrast Normalization (GCN), as Equation~\ref{eq:5} shows.

\begin{equation} \label{eq:5}
\mathsf{X}^{\prime}_{i,j,k}=s\frac{\mathsf{X}_{i,j,k}-\overline{\mathsf{X}}}{max\left\lbrace \epsilon, \sqrt{\alpha+\frac{1}{3rc}\sum_{i=1}^{r}\sum_{j=1}^{c}\sum_{k=1}^{3}(\mathsf{X}_{i,j,k}-\overline{\mathsf{X}})^2}\right\rbrace }
\end{equation}

where, $X$ and $X^{'}$ are the raw image and the normalized image. $s$, $\epsilon$ and $\alpha$ are meta-parameters whose values are 1, $1e^{-9}$, and 10. Then we apply Zero Component Analysis (ZCA) to the whole dataset. Specifically, we choose 10000 images $\mathsf{X^{''}}$ randomly from the GCN-processed training set and calculate the mean image $\overline{\mathsf{X^{''}}}$ across all the pixels. Then we calculate the covariance matrix as well as the singular values and vectors, as, Equation~\ref{eq:6} shows.

\begin{equation} \label{eq:6}
\mathsf{U}, \mathsf{S}, \mathsf{V} = SVD(\Cov(\mathsf{X^{''}} - \overline{\mathsf{X^{''}}})) 
\end{equation}

Finally, we can use Equation~\ref{eq:7} to process each image in the dataset.
\begin{equation} \label{eq:7}
\mathsf{X_{ZCL}} = \mathsf{U} \ \displaystyle \text{diag}(\frac{1}{ \sqrt{ \displaystyle \text{diag}(\mathsf{S})} + 0.1}) \mathsf{U^T}
\end{equation}

\begin {table}[h!]
\begin{center} 
\begin{tabular}{ lcc } 
\toprule
Batch Size & \textbf{CPU}(s/100 iterations) & \textbf{CUDA Kernel}(s/100 iterations) \\ 
\midrule
50 & 106.22 & 19.67\\
100 & 213.60 & 46.57\\
150 & 319.37 & 61.63\\ 
200 & 425.15 & 91.59\\
\bottomrule
\end{tabular}
\caption{Comparison on time consumed between CPU mode and acceleration solution}
\label{tab:runtime} 
\end{center}
\end{table}

\section{Acceleration Solution for P-CapsNets}
\label{app: imple}
Different from convolution operations in CNNs, which can be interpreted as a few large matrix multiplications during training, the capsule convolutions in P-CaspNets have to be interpreted as a large number of small matrix multiplication. If we use the current acceleration library like CuDNN~(\cite{cuDNN}) or the customized convolution solution in CAFFE~(\cite{caffe}), too many communication times would be incorporated which slows the whole training process a lot. The communication overhead is so much that the training is slower than CPU-only mode. To overcome this issue, we parallel the operations within each kernel to minimize communication times. We build two P-CaspNets\#3 models, one is CPU-only based, the other one is based on our parallel solution. The GPU is one TITAN Xp card, the CPU is Intel Xeon. As Table~\ref{tab:runtime} shows, our solution achieves at least $4\times$ faster speed than the CPU mode for different batch sizes.

% {\small
% \bibliographystyle{ieee_fullname}
% \bibliography{egbib}
% }

\end{document}